%% file: arxiv_2020_reservoir_memory_machines.tex
\documentclass[a4paper,11pt]{article}

\usepackage{amssymb,amsmath,array,bm}
\usepackage{booktabs}

\usepackage{tikz}
\usetikzlibrary{arrows}

\usepackage{pgfplots}
\pgfplotsset{compat=1.13}
\usepgfplotslibrary{groupplots}

\usepackage{hyperref}
\usepackage[utf8]{inputenc}
\usepackage[T1]{fontenc}
\usepackage[english]{babel}

\newcommand{\N}{\mathbb{N}}
\newcommand{\R}{\mathbb{R}}

\usepackage{tango-colors}
\input{tikz_styles.tex}

\usepackage[a4paper, left=3cm, right=2.5cm, top=3cm, bottom=3cm]{geometry}

\usepackage{authblk}
\usepackage{natbib}
\usepackage{doi}

\usepackage[utf8]{inputenc}
\usepackage[T1]{fontenc}
\usepackage[english]{babel}
\usepackage{csquotes}

\renewcommand{\cite}{\citep}

\author{Benjamin Paaßen}
\author{Alexander Schulz}
\affil{Machine Learning Group, Bielefeld University\thanks{Funding by the Bielefeld Young Researchers' Fund and from BMBF within the project MechML under grant number 01IS18053E is gratefully acknowledged. We also thank Barbara Hammer for brilliant theoretical insights.}}

\title{Reservoir memory machines}

\date{Preprint of the ESANN 2020 paper \citet{Paassen2020ESANN} as provided by the authors.
The original can be found at the \href{https://www.esann.org/proceedings/2020}{ESANN electronics proceedings page}.}

\begin{document}

\maketitle

\pagestyle{myheadings}
\markright{Preprint of \citet{Paassen2020ESANN} provided by the authors.}

\input{paassen_schulz_RMM_content}

\bibliography{literature} 
\bibliographystyle{plainnat}

\end{document}

%% file: tikz_styles.tex
\tikzstyle{neuron}=[circle, draw=black, inner sep=0pt, minimum size=2mm, semithick, anchor=center]
\tikzstyle{edge}=[->, >=stealth', shorten <=2pt, shorten >=2pt, auto, semithick]
\tikzstyle{curves}=[line width=0.5mm]
\tikzstyle{curves_thin}=[line width=0.25mm]
\tikzstyle{curves_thick}=[line width=0.75mm]
\tikzstyle{class0color}=[aluminium6]
\tikzstyle{class0}=[draw=aluminium6, fill=aluminium4, text=aluminium6]
\tikzstyle{class1color}=[skyblue3]
\tikzstyle{class1}=[draw=skyblue3, fill=skyblue1, text=skyblue3]
\tikzstyle{class2color}=[orange3]
\tikzstyle{class2}=[draw=orange3, fill=orange1, text=orange3]
\tikzstyle{class3color}=[chameleon3]
\tikzstyle{class3}=[draw=chameleon3, fill=orange1, text=chameleon3]

%% file: paassen_schulz_RMM_content.tex
\begin{abstract}
In recent years, Neural Turing Machines have gathered attention by joining
the flexibility of neural networks with the computational capabilities of Turing
machines. However, Neural Turing Machines are notoriously hard to train,
which limits their applicability. We propose reservoir memory machines, which
are still able to solve some of the benchmark tests for Neural Turing Machines,
but are much faster to train, requiring only an alignment algorithm and linear
regression. Our model can also be seen as an extension of echo state
networks with an external memory, enabling arbitrarily long storage without
interference.
\end{abstract}

\section{Introduction}

While neural networks have achieved impressive successes in domains like image
classification or machine translation, standard models still struggle with tasks
that require very long-term memory without interference and would thus benefit
from a separation of memory and computation \cite{NTM,NTM_impl}.
Neural Turing Machines (NTM) attempt to address these tasks by augmenting recurrent neural
networks with an explicit memory to which the network has read and write
access \cite{NTM,NTM_impl}. Unfortunately, such models are notoriously hard to train,
even compared to other deep learning models \cite{NTM_impl}.

In our contribution, we propose to address this training problem by replacing
the learned recurrent neural network controller of a NTM with
an echo state network (ESN) \cite{ESN}. In other words, we only learn the
controller for the read and write head of our memory access as well as the output
mapping, all of which is possible via standard linear regression. To construct
the training data for our read and write head controllers, we only require a
standard dynamic time warping alignment. We call this model a \emph{reservoir memory machine} (RMM).

Our model can also be seen as an augmentation of echo state networks with
an explicit external memory, such that input information can be stored
for arbitarily long times without interference, whereas the maximum memory
horizon for regular echo state networks is limited to the number of neurons in
the reservoir \cite{ESN,MemCap,DeepESN}.

In the remainder of this paper, we first refresh the reader's memory regarding
standard ESNs, then formally define our own model - reservoir memory machines -,
and finally show that our proposed model is sufficient to solve three
benchmark tasks for Neural Turing Machines with much faster training.

\section{Echo state networks}

An echo state network (ESN) \cite{ESN} is a recurrent network, i.e.\
the neural activations $\vec h_t \in \R^m$ at time $t$ are computed as $\vec h_t
= \mathrm{tanh}\Big( \bm{U} \cdot \vec x_t + \bm{W} \cdot \vec h_{t-1} \Big)$,
where $\vec x_t \in \R^n$ is the input at time $t$, $\bm{U} \in \R^{m \times n}$
are the input weights, and $\bm{W} \in \R^{m \times m}$ are the recurrent weights
of the network. The output $\vec y_t \in \R^L$ of the network at time $t$ is
computed as $\vec y_t = \bm{V} \cdot \vec h_t$, where $\bm{V} \in \R^{L \times m}$
are the output weights. ESNs have two distinct characteristics.
First, $\bm{U}$ and $\bm{W}$ are not learned but kept fixed after
initialization. This means that the activations $\vec h_1, \ldots, \vec h_T$
can be seen as a nonlinear preprocessing of the input, which makes learning $\bm{V}$
a generalized linear regression problem that can be solved analytically with the
pseudo-inverse.
Second, the recurrent weights $\bm{W}$ must ensure the \emph{echo state property},
i.e.\ past influences must degrade over time \cite{ESN,CRJ}.
This property is necessary to ensure that the network's dynamic is independent of
initial conditions and always adjusts to the input time series. On the other hand,
it necessarily limits ESNs to \emph{short term} memory tasks. In particular the
memory is upper-bounded by the number of neurons $n$ \cite{ESN,MemCap}. This is the
key limitation we aim to address.

In this paper, we employ the deterministic
'cycle reservoir with jumps' scheme to initialize $\bm{U}$
and $\bm{W}$ \cite{CRJ}. In this scheme, the entries of $\bm{U}$ are set to a constant value
$u \in (-1, 1)$ with a sign determined by a fixed, aperiodic sequence
(e.g.\ the digits of pi), and $\bm{W}$ is a sparse matrix with off-diagonal
cycle connections $w_{i, i+1} = w_c \in [0, 1)$ and longer 'jump' connections
$w_{i, i+l} = w_{i+l, i} = w_l \in [0, 1)$. Note that $u$, $w_c$, $w_j$, and
$l \in \N$ are hyper-parameters of the model. Because this initialization is deterministic,
we can compare different architectures more easily. In general, however, our architecture
is agnostic regarding the initialization.

\section{Reservoir memory machines}

\begin{figure}
\begin{center}
\begin{tikzpicture}
\node at (0,0.5) {input};
\node (x) at (0,0) {$\vec x_t$};

\begin{scope}[shift={(2.5,0)}]

\node[neuron] (h1) at (180:0.6) {};
\node[neuron] (h2) at (252:0.6) {};
\node[neuron] (h3) at (324:0.6) {};
\node[neuron] (h4) at (036:0.6) {};
\node[neuron] (h5) at (108:0.6) {};

\node[left] at (0,1) {reservoir};

\path[edge]
(h1) edge[bend right] (h2)
(h2) edge[bend right] (h3)
(h3) edge[bend right] (h4)
(h4) edge[bend right] (h5)
(h5) edge[bend right] (h1)
(h1) edge[<->] (h3)
(h3) edge[<->] (h5);

\node at (0,0) {$\bm{W}$};

\node (h) at (1.1,0) {$\vec h_t$};
\end{scope}

\path[edge]
(x) edge node[above] {$\bm{U}$} (h1);

\begin{scope}[shift={(5.5,0)}]
\node[right] at (1.4,0.4) {memory};

\draw[semithick] (0,-0.25) rectangle (2.25,+0.25);
\draw[semithick] (0.75,-0.25) -- (0.75,+0.25);
\draw[semithick] (1.5,-0.25) -- (1.5,+0.25);

\node at (0.4,0) {$\vec m_{t, 1}$};
\node at (1.15,0) {$\ldots$};
\node at (1.9,0) {$\vec m_{t, K}$};

\node[right] at (1.35,0.9) {$\strut$write head};
\node[rectangle, draw=black] (write) at (1.15,0.9) {$\vec x_t$};
\node (k) at (1.15,0.15) {};
\path[edge, shorten <=0pt] (write) edge (k);
\path[edge, dashed, gray]
(x) edge[out=30,in=180, looseness=0.75] (write)
(h) edge (write);
\node[gray] at (+0.4,1.1) {$\vec u^w$};
\node[gray] at (+0.5,0.5) {$\vec v^w$};

\node[right] at (1.35,-0.9) {$\strut$read head};
\node[rectangle, draw=black] (read) at (+1.15,-0.9) {$\vec r_t$};
\node (l) at (1.15,-0.15) {};
\path[edge, shorten >=0pt] (l) edge (read);
\path[edge, dashed, gray]
(x) edge[out=-30,in=180, looseness=0.75] (read)
(h) edge (read);
\node[gray] at (+0.4,-1.1) {$\bm{U}^r$};
\node[gray] at (+0.65,-0.58) {$\bm{V}^r$};
\end{scope}

\node at (10,0.5) {output};
\node (y) at (10,0) {$\vec y_t$};
\path[edge]
(h) edge[out=-15,in=190, looseness=0.8] (y)
(read) edge (y);
\node at (9.3,0.1) {$\bm{V}$};
\node at (9.5,-0.4) {$\bm{R}$};
\end{tikzpicture}

\vspace{-0.5cm}
\end{center}
\caption{An illustration of reservoir memory machines. We first process the
input (left) with a cycle reservoir with jumps (center left). We then use input and
reservoir activations to control interaction with the memory (center right;
gray connections). Finally, we feed reservoir activations and memory
reads to the output (right).}
\label{fig:rmm}
\end{figure}
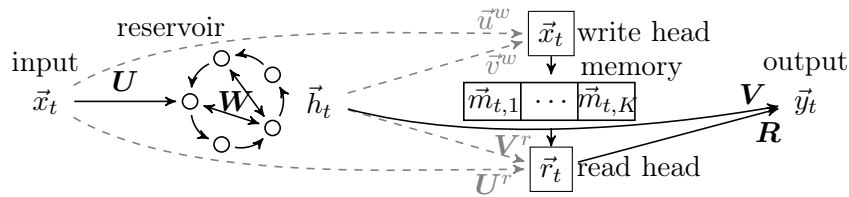

Our key contribution is an easy-to-train alternative to the Neural Turing Machine \cite{NTM}.
In particular, we propose to extend an ESN with an explicit memory,
a write head, which can copy inputs into memory, and a read head, which can
read from the memory. We call this augmented ESN version
a \emph{reservoir memory machine} (RMM). A sketch of the RMM architecture
is shown in Figure~\ref{fig:rmm}.

In more detail, the state of our system is now a quadruple $(\vec h_t, \bm{M}_t, k_t, l_t)$,
where $\vec h_t$ are the reservoir activations as before, $\bm{M}_t \in \R^{K \times n}$
is the current memory state (of size $K$), and $k_t, l_t \in \{1, \ldots, K\}$ are the
current position of the write and read head respectively.

The dynamics of the system are as follows. First, we copy the previous
memory state, i.e.\ $\bm{M}_t \gets \bm{M}_{t-1}$ (where $\bm{M}_{-1} = \bm{0}$).
Then, we control the write head with the value $c^w_t = \vec u^w \cdot \vec x^t + \vec v^r \cdot \vec h_t$,
where $\vec u^w \in \R^n$ and $\vec v^r \in \R^m$ are learnable parameters.
If $c^w_t > 0$, we write to the memory, i.e.\ $\vec m_{t, k} \gets \vec x_t$,
and increment $k_t \gets k_{t-1} + 1$ (re-setting $k_t$ to $1$ if it exceeds $K$).
Otherwise, we leave the memory and $k_t$ as is.
Similarly, in each time step we control the read head with the vector
$\vec c^r_t = \bm{U}^r \cdot \vec x_t + \bm{V}^r \cdot \vec h_t$, where
$\bm{U}^r \in \R^{3 \times n}$ and $\bm{V}^r \in \R^{3 \times m}$ are learnable parameters.
If $c^r_{t, 1} = \max\{ c^r_{t, 1}, c^r_{t, 2}, c^r_{t, 3} \} $, the read head stays in the same location, i.e.\
$l_t \gets l_{t-1}$; if $c^r_{t, 2} = \max\{ c^r_{t, 1}, c^r_{t, 2}, c^r_{t, 3} \}$, we increment $l_t \gets l_{t-1} + 1$
(re-setting $l_t$ to 1 if it exceeds $K$); otherwise, we re-set $l_t \gets 1$.
We then set the memory read at time $t$ as the $l_t$th row of $\bm{M}_t$, i.e.\
$\vec r_t \gets \vec m_{t, l_t}$.

The output of the system at time $t$ is $\vec y_t = \bm{V} \cdot
\vec h_t + \bm{R} \cdot \vec r_t$, where $\bm{V} \in \R^{L \times m}$ and $\bm{R} \in \R^{L \times n}$
are learnable parameters. Note that our proposed model is a strict extension
of an ESN because we can simply set $\bm{R} = \bm{0}$ and thus obtain a standard ESN.
However, we can potentially solve \emph{more} tasks.

\paragraph{Training:} Because the output generation depends on the memory content,
our first step is to train the write and read heads, i.e.\ the
parameters $\vec u^w$, $\vec v^w$, $\bm{U}^r$, and $\bm{V}^r$.
In more detail, we initialize
$\bm{R}$ as the identity matrix (padded with zeros whenever necessary) and then
identify for each output $\vec y_t$ the earliest input $\vec x_{\tau_t}$ that
minimizes the distance $\lVert \bm{R} \cdot \vec x_{\tau_t} - \vec y_t \rVert$.
Based on this, we generate an \emph{ideal} control sequence for the write head
$c^w_1, \ldots, c^w_T$ where $c^w_t = +1$ if $t \in \{\tau_1, \ldots, \tau_T\}$
and $c^w_t = -1$ otherwise.
This control sequence serves as our teaching signal for training
$\vec u^w$ and $\vec v^w$ via linear regression.

Next, we generate the tensor of all memory states
$(\bm{M}_1, \ldots, \bm{M}_T) \in \R^{T \times K \times n}$ as described above.
We then align this tensor with the output time series $\vec y_1, \ldots, \vec y_T$
via a variant of dynamic time warping with the recurrence:
$d_{l, t} = \lVert \bm{R} \cdot \vec m_{t, l} - \vec y_t \rVert +
\min\{ d_{l, t+1}, d_{l+1, t+1}, d_{1, t+1} \}$, where the entries in the minimum
correspond respectively to leaving the read-head location as is, incrementing it,
or resetting it to one. The base case of this recurrence is
$d_{l, T+1} = 0$ for all $l \in \{1, \ldots, K\}$. 
Note that $\min\{d_{1, 1}, d_{2, 1}\}$ then corresponds to the error we achieve by
optimally moving the read head over the memory and always predicting the output
$\bm{R} \cdot m_{t, l_t}$. Accordingly, backtracing yields a teaching signal
to train the read head parameters $\bm{U}^r$ and $\bm{V}^r$ via linear regression.

Finally, we compute the sequence of memory reads $\vec r_1, \ldots, \vec r_T$ as
described above, which we use to train both
$\bm{V}$ and $\bm{R}$ via linear regression. Now, because we change $\bm{R}$,
the optimal alignments in the previous steps may change as well. Accordingly,
we repeat the training process until the loss increases or until
convergence, yielding an alternating optimization algorithm. 

\section{Experiments}

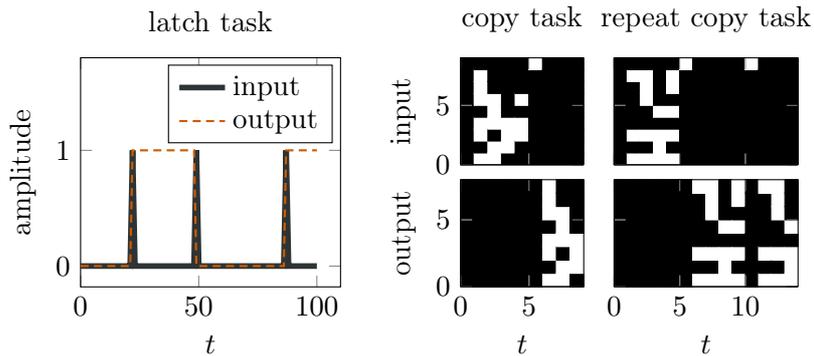
\begin{figure}
\begin{center}
\begin{tikzpicture}
\begin{scope}[shift={(-5,-1.62)}]
\begin{axis}[title={latch task},xlabel={$t$},ylabel={amplitude}, width=5cm, height=4.62cm,
xmin=0, ymax=1.8, legend pos={north east}, legend cell align={left},
ytick={0,1}]
\addplot[class0color, curves_thick] table[x=time,y=x] {latch_example.csv};
\addlegendentry{input}
\addplot[thick, class2color, densely dashed] table[x=time,y=y] {latch_example.csv};
\addlegendentry{output}
\end{axis}
\end{scope}
\begin{groupplot}[view={0}{90}, xlabel={$t$}, ymin=0, ymax=8,
group style={group size=2 by 2,
  x descriptions at=edge bottom,y descriptions at=edge left,
  horizontal sep=0.4cm, vertical sep=0.2cm},
width=4cm, height=3cm,
colormap/blackwhite]
\nextgroupplot[title={copy task},ymax=9,width=3.2cm, ylabel={input}]
\addplot3[surf,mesh/ordering=y varies,mesh/rows=10,shader=flat corner] file {copy_example_input.csv};
\nextgroupplot[title={repeat copy task},ymax=9]
\addplot3[surf,mesh/ordering=y varies,mesh/rows=10,shader=flat corner] file {repeat_copy_example_input.csv};
\nextgroupplot[width=3.2cm, ylabel={output}]
\addplot3[surf,mesh/ordering=y varies,mesh/rows=9,shader=flat corner] file {copy_example_output.csv};
\nextgroupplot
\addplot3[surf,mesh/ordering=y varies,mesh/rows=9,shader=flat corner] file {repeat_copy_example_output.csv};
\end{groupplot}
\end{tikzpicture}
\vspace{-0.7cm}
\end{center}
\caption{An example input and output sequence for all three data sets.}
\label{fig:data}
\end{figure}

In our experiments, we evaluate reservoir memory machines (RMMs) on three data sets that
require storage of inputs over long times without interference:

The \textit{latch} task requires to produce zeros until a spike in the input appears,
after which the model should produce ones. For the next input spike, the model should switch back to zeros,
and so on (Figure~\ref{fig:data}, left). We use three spikes with random positions and random sequence lengths of up to 200 time steps.

The \textit{copy} data set \cite{NTM} consists of 1-20 time steps with 8 random bits each, followed by a sequence
end token in an additional feature. After this, the goal is to exactly copy the input while the remaining input is
zero (Figure~\ref{fig:data}, center).

The \textit{repeat copy} data set \cite{NTM} extends the copy task by requiring the network to
copy the input sequence multiple times (refer to Figure~\ref{fig:data}, right).

We compare RMMs to standard ESNs and to a novel variant which we dub echo state
gated recurrent unit (ESGRU). This model uses the dynamic equations of a gated
recurrent units \cite{GRU} but keeps all weights fixed after initialization.
To ensure that all variance is due to memory access only, we use the same
reservoir for all networks, namely a cycle reservoir with jumps \cite{CRJ}.

We evaluate in a 20 fold crossvalidation, generating
10 sequences per fold (i.e.\ $190$ training sequences and $10$ test sequences). For each model,
we used a 3-fold nested crossvalidation for hyper-parameter optimization via random
search with 10 trials. The detailed experimental code is available at \url{https://gitlab.ub.uni-bielefeld.de/bpaassen/reservoir-memory-machines}.

\begin{table}
\caption{The average RMSE ($\pm$ standard deviation) across 20 crossvalidation
folds for all models and all data sets. NTM results are copied from \cite{NTM_impl}.}
\label{tab:results}
\begin{center}
\begin{tabular}{lccc}
\textbf{model} & \textbf{latch} & \textbf{copy} & \textbf{repeat copy} \\
\cmidrule(lr){1-1}\cmidrule(lr){2-2}\cmidrule(lr){3-3}\cmidrule(lr){4-4}
ESN & $0.309 \pm 0.049$ & $0.358 \pm 0.030$ & $0.409 \pm 0.038$ \\
ESGRU & $0.402 \pm 0.116$ & $0.331 \pm 0.011$ & $0.375 \pm 0.018$ \\
RMM & $<10^{-3}$ & $0.027 \pm 0.025$ & $0.037 \pm 0.067$ \\
NTM & n.a. & $ < 10^{-3} $ \cite{NTM_impl} & $ < 10^{-3} $ \cite{NTM_impl}
\end{tabular}
\end{center}
\end{table}

The generalization root mean square error (RMSE) of all models on all datasets
is displayed in Table~\ref{tab:results}.
For all datasets, RMMs achieve a low (albeit nonzero) error, indicating that RMMs
are able to solve the tasks.
Additionally, we note that both ESNs and ESGRUs are \emph{not} able to solve the
tasks, because they have significantly higher errors in all datasets
($p < 10^{-3}$ according to a Wilcoxon sign-rank test with Bonferroni correction).
Note that a Neural Turing Machine achieves zero error on all tasks \cite{NTM_impl}.

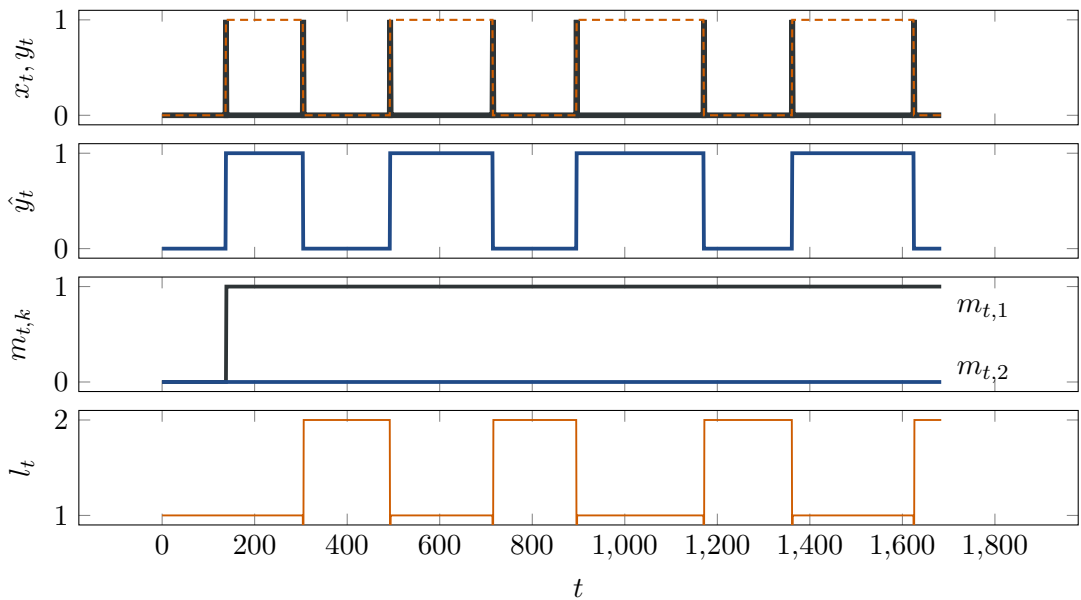
\begin{figure}
\begin{center}
\begin{tikzpicture}
\begin{groupplot}[
    group style={
        group size=1 by 4, horizontal sep=0.5cm, vertical sep=0.25cm,
        x descriptions at=edge bottom,
        y descriptions at=edge left,
    },
    width = 0.95\textwidth,
    height = 0.2\textwidth,
    enlarge x limits,
    enlarge y limits,
    xmin=0, xmax=1800
]

\nextgroupplot[ymin=0, ymax=1, ylabel={$x_t,y_t$}, ytick={0, 1}]
\addplot[class0color, curves_thick] table[x=time, y=x] {latch_example_signal.csv};
\addplot[class2color, thick, densely dashed] table[x=time, y=y] {latch_example_signal.csv};

\nextgroupplot[ymin=0, ymax=1, ylabel={$\hat{y_t}$}, ytick={0, 1}] 
\addplot[class1color, curves] table[x=time, y=yhat] {latch_example_signal.csv};

\nextgroupplot[ymin=0, ymax=1, ylabel={$m_{t, k}$}, ytick={0, 1}] 
\addplot[class0color, curves] table[x=time, y=mem1] {latch_example_signal_writesCont.csv};
\addplot[class1color, curves] table[x=time, y=mem2] {latch_example_signal_writesCont.csv};
\node (m1) at (1775,0.75) {$m_{t, 1}$};
\node (m2) at (1775,0.1) {$m_{t, 2}$};

\nextgroupplot[ymin=1, ymax=2, xlabel={$t$}, ylabel={$l_t$}, ytick={1, 2}] 
\addplot[class2color, curves_thin] table[x=time, y=read_loc] {latch_example_signal_readsCont.csv};

\end{groupplot}
\end{tikzpicture}
\vspace{-0.5cm}
\end{center}
\caption{From top to bottom: A long sequence from the latch task with the input as solid, the output as dashed line; 
the prediction of the RMM;
the memoy entries over time; and
the read head position over time.}
\label{fig:latch_example}
\end{figure}

We investigate the solution strategy of the RMM model in more detail on the
latch task. For this purpose, we use a trained RMM and let it extrapolate
to a much longer sequence (see Figure~\ref{fig:latch_example}, top)
than seen in training (length 1700 vs.\ 200 with 8 vs.\ 3 spikes). We note that the RMM extrapolates
perfectly (Figure~\ref{fig:latch_example}, second row) with an error $<10^{-3}$.
In more detail, we observe that the model only writes to memory once, namely storing
a 1 at the time of the first spike (Figure~\ref{fig:latch_example}, third row),
whereas the read head switches position at every spike (except the first one;
Figure~\ref{fig:latch_example}, bottom), thus producing the desired output.

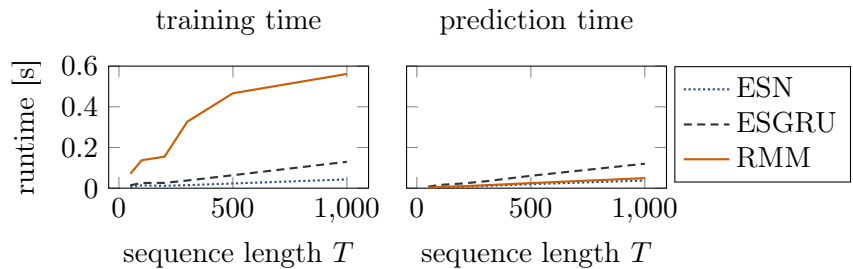
\begin{figure}
\begin{center}
\begin{tikzpicture}
\begin{groupplot}[xlabel={sequence length $T$},
group style={group size=2 by 1, horizontal sep=0.5cm, y descriptions at=edge left},
width=5cm, height=3.2cm,ymin=0,ymax=0.6]
\nextgroupplot[title={training time$\strut$}, ylabel={runtime [s]}]
\addplot[thick, class1color, densely dotted]
table[x=length,y=ESN_train_mean,y error=ESN_train_std] {runtimes.csv};
\addplot[thick, class0color, densely dashed]
table[x=length,y=ESGRU_train_mean,y error=ESGRU_train_std] {runtimes.csv};
\addplot[thick, class2color]
table[x=length,y=ARMM_train_mean,y error=ARMM_train_std] {runtimes.csv};
\nextgroupplot[title={prediction time$\strut$}, legend pos={outer north east}, legend cell align={left}]
\addplot[thick, class1color, densely dotted]
table[x=length,y=ESN_pred_mean,y error=ESN_pred_std] {runtimes.csv};
\addlegendentry{ESN}
\addplot[thick, class0color, densely dashed]
table[x=length,y=ESGRU_pred_mean,y error=ESGRU_pred_std] {runtimes.csv};
\addlegendentry{ESGRU}
\addplot[thick, class2color]
table[x=length,y=ARMM_pred_mean,y error=ARMM_pred_std] {runtimes.csv};
\addlegendentry{RMM}
\end{groupplot}
\end{tikzpicture}
\vspace{-0.7cm}
\end{center}
\caption{Runtime results for training (left) and prediction (right)
of standard ESNs, ESGRUs, and RMMs for varying sequence length.}
\label{fig:runtimes}
\end{figure}

To evaluate the runtime, we train ESNs, ESGRUs, and RMMs
with a reservoir of 128 neurons each on a random 8-bit input sequence
with varying length, the output sequence being shifted by one.
We measure runtime on a consumer grade laptop with core i7 CPU.
Figure~\ref{fig:runtimes} shows the runtime results. We find that RMMs roughly
take 15 times longer to train compared to regular ESNs, which may be due to
more needed linear regression runs and an inefficient alignment implementation.
Still, even for long sequences we maintain training times well below a second.
Prediction time is roughly comparable to a standard ESN and faster than an ESGRU.
By comparison, training a NTM using the reference implementation
\cite{NTM_impl} on the copy task took more than 30 minutes.

\section{Conclusion}

We have introduced reservoir memory machines (RMMs), which augment echo state
networks with an external memory, a write head that copies data from the
input to the memory, and a read head which couples the memory to the
output. We also provided a training algorithm for the write and read heads
based on dynamic time warping and linear regression in
an alternating optimization scheme. As such, our model retains the
training simplicity of echo state networks, but extends
its capabilities to some of the benchmark tasks of Neural Turing Machines.
We emphasize that our model is still strictly less powerful because other
benchmark tasks remain out of reach, especially those based on
content-based addressing.
Extending our model with such a mechanism is a task for future work.
Further, we still require a formal proof that our proposed model is
strictly more powerful than an ESN.